\documentclass{article}

\usepackage{microtype}
\usepackage{graphicx}
\usepackage{booktabs} % for professional tables
\usepackage{xcolor}
\usepackage{graphicx}
\usepackage{subcaption}
\usepackage{appendix}
\usepackage{resizegather}
\usepackage{multirow}
\usepackage{hyperref}

\usepackage[accepted]{icml2021}

\icmltitlerunning{In-Time Over-Parameterization}

\begin{document}

\twocolumn[
\icmltitle{Do We Actually Need Dense Over-Parameterization?\\ In-Time Over-Parameterization in Sparse Training}
% \icmltitle{Beyond Over-Parameterization:\\ In-Time Over-Parameterization in Sparse Training}




\begin{icmlauthorlist}
\icmlauthor{Shiwei Liu}{tue}
\icmlauthor{Lu Yin}{tue}
\icmlauthor{Decebal Constantin Mocanu}{tue,twente}
\icmlauthor{Mykola Pechenizkiy}{tue}
\end{icmlauthorlist}

\icmlaffiliation{tue}{Department of Mathematics and Computer Science,
              Eindhoven University of Technology, 5600 MB Eindhoven, the Netherlands\\}
\icmlaffiliation{twente}{Faculty of Electrical Engineering, Mathematics andComputer Science, University of Twente, Enschede 7522NB,The Netherlands\\}

\icmlcorrespondingauthor{Shiwei Liu}{s.liu3@tue.nl}
\icmlkeywords{Machine Learning, ICML}

\vskip 0.3in
]



\printAffiliationsAndNotice{} % otherwise use the standard text.

\begin{abstract}
% Despite the dominating performance of the advanced deep learning models, the ever-increasing size of these highly over-parameteried models require prohibitive costs to train and deploy them.
In this paper, we introduce a new perspective on training deep neural networks capable of state-of-the-art performance without the need for the expensive over-parameterization by proposing the concept of In-Time Over-Parameterization (ITOP) in sparse training. By starting from a random sparse network and continuously exploring sparse connectivities during training, we can perform an Over-Parameterization over the course of training, closing the gap in the expressibility between sparse training and dense training. We further use ITOP to understand the underlying mechanism of Dynamic Sparse Training (DST) and discover that the benefits of DST come from its ability to consider across time all possible parameters when searching for the optimal sparse connectivity. As long as sufficient parameters have been reliably explored, DST can outperform the dense neural network by a large margin. We present a series of experiments to support our conjecture and achieve the state-of-the-art sparse training performance with ResNet-50 on ImageNet. More impressively, ITOP achieves dominant performance over the overparameterization-based sparse methods at extreme sparsities. When trained with ResNet-34 on CIFAR-100, ITOP can match the performance of the dense model at an extreme sparsity of 98\%. 
% \textcolor{blue}{In addition, we observe the ability of In-Time Over-Parameterization to improve generalization.} 

\end{abstract}

\section{Introduction}
\label{submission}
\begin{figure}[ht]
\vspace{-0.3cm}
        \centering
        \includegraphics[width=0.45\textwidth]{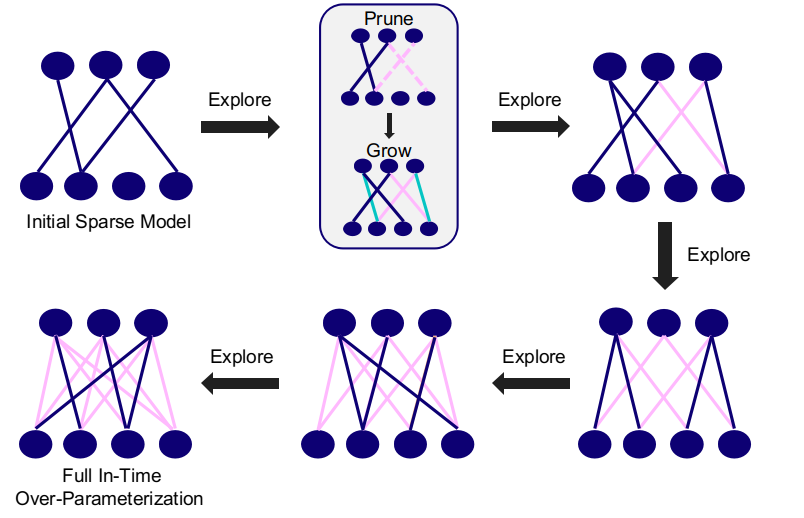}
        \caption[ The average and standard deviation of critical parameters ]
        {\small As the figure proceeds, we perform an Over-Parameterization in time. Blue lines refer to the currently activated connections. Pink lines are the connections that have been activated previously. While exploring In-Time Over-Parameterization, the parameter count (blue lines) of the sparse model is fixed throughout training.}
        \label{fig:ITOP}
        \vspace{-0.5cm}
    \end{figure}
Over-Parameterization has been shown to be crucial to the dominating performance of deep neural networks in practice, despite the fact that the training objective function is usually non-convex and non-smooth \citep{goodfellow2014qualitatively,brutzkus2017sgd,li2018learning,safran2018spurious,soudry2016no,allen2019convergence,du2019gradient,zou2020gradient,zou2019improved}. Meanwhile, advanced deep models \citep{simonyan2014very,he2016deep,devlin2018bert,brown2020language,dosovitskiy2021an} are continuously achieving state-of-the-art results in numerous machine-learning tasks. While achieving impressive performance, the size of the state-of-the-art models is also exploding. The resources required to train and deploy those highly over-parameterized models are prohibitive.
\begin{figure*}[t]
\vspace{-0.1em}
    \centering
    \includegraphics[width=\textwidth]{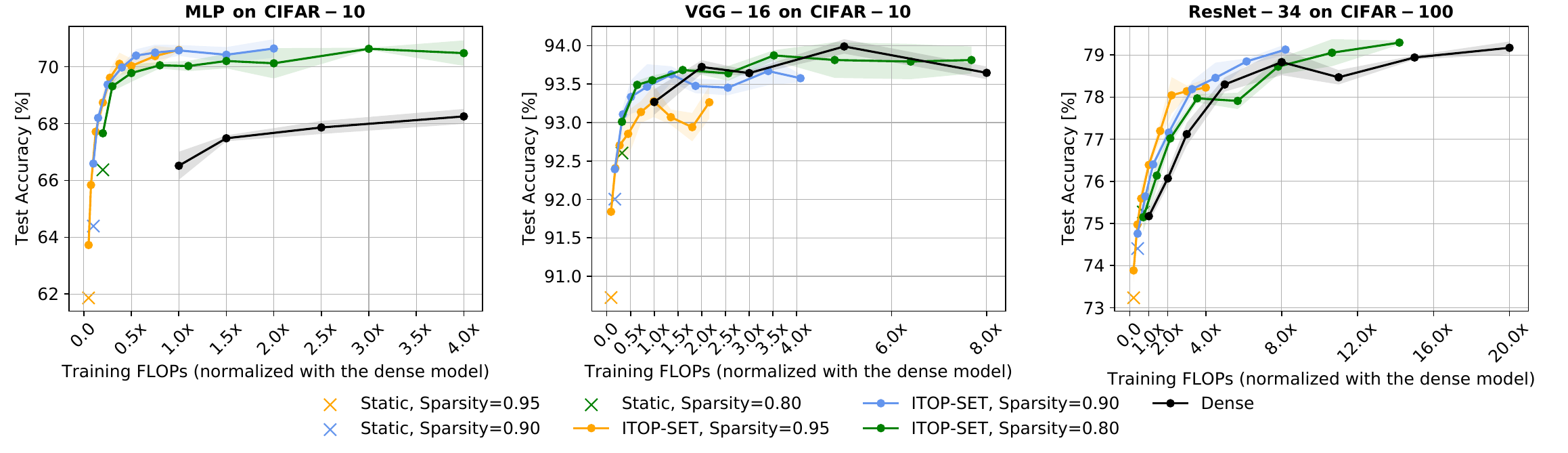}
    \vspace{-0.8cm}
    \caption
        {\small Training FLOPs of sparse models trained with In-Time Over-Parameterization. $\Delta T=1500$ and batch size is 128. }  
    \label{Fig:flops}
    \vspace{-0.3em}
\end{figure*}

Motivated for inference, a large body of research \citep{mozer1989using,han2015learning} attempts to discover a sparse model that can sufficiently match the performance of the corresponding dense model while substantially reduce the number of parameters. While effective, these techniques involve pre-training a highly over-parameterized model for either at least a full converged training time (full dense over-parameterization) \citep{janowsky1989pruning,lecun1990optimal,hassibi1993second,molchanov2017variational,han2015deep,gomez2019learning,dai2018compressing} or a partial converged training time (partial dense over-parameterization) \citep{louizos2017learning,zhu2017prune,gale2019state,savarese2019winning,kusupati2020soft,you2019drawing}. Given the fact that the training costs of state-of-the-art models e.g., GPT-3 \citep{brown2020language} and Vision Transformer \citep{dosovitskiy2020image}, have increasingly exploded, this heavily over-parameterized dependency leads to a situation where state-of-the-art models are beyond the reach of the majority of the machine learning community.

Recently, the lottery ticket hypothesis (LTH) \citep{frankle2018the} shows the possibility to train a sub-network from scratch (sparse training) to match the performance of the dense network. However, these ``winning tickets'' are found under the guidance of a fully dense over-parameterized process (iterative pruning a fully converged network), and solutions that are discovered through either a partial dense over-parameterization (pruning at initialization \citep{lee2018snip,Wang2020Picking,de2020progressive}) or no over-parameterization (randomly-initialized static sparse training \citep{mocanu2016topological,evci2019difficulty}) typically are not able to match the accuracy achieved by their dense counterpart. A common-sense explanation would be that in comparison with dense training, sparse training, especially at extreme high sparsities, does not have the over-parameterization property, and hence suffers from a poor expressibility. One approach to address this problem is to leverage the knowledge learned from dense training, e.g., LTH \citep{frankle2018the}. While effective, the computational costs and memory requirements attached to the over-parameterized dense training are prohibitive. 

\subsection{Our Contribution}
In this paper, we propose a concept that we call \textit{In-Time Over-Parameterization}\footnote{\url{https://github.com/Shiweiliuiiiiiii/In-Time-Over-Parameterization}} to close the gap in over-parameterization along with expressibility between sparse training and dense training, illustrated in Figure \ref{fig:ITOP}. Instead of inheriting weights from a dense and pre-trained model, allowing a continuous parameter exploration across the training time performs an over-parameterization in the space-time manifold, which can significantly improve the expressibility of sparse training. 

We find the concept of In-Time Over-Parameterization useful (1) in exploring the expressibility of sparse training, especially for extreme sparsities, (2) in reducing training and inference costs (3) in understanding the underlying mechanism of dynamic sparse training (DST) \citep{mocanu2018scalable,evci2019rigging}, (4) in preventing overfitting and improving generalization.

Based on In-Time Over-Parameterization, we improve the state-of-the-art sparse training performance with ResNet-50 on ImageNet. We further assess the ITOP concept by applying it to the main class of sparse training methods, DST, in comparison with the overparameterization-based sparse methods including LTH, gradual magnitude pruning (GMP), and pruning at initialization (PI). Our results show that, when a sufficient and reliable parameter exploration is reached (as required by ITOP), DST consistently outperforms those overparameterization-based methods. Since ITOP eliminates the dense Over-Parameterization throughout the whole course of training, it can match the performance of the corresponding dense networks with much fewer training FLOPs, as shown in Figure~\ref{Fig:flops}.

% We further compare the In-Time Over-Parameterization concept applied to the main class of sparse training methods, DST, against the overparameterization-based sparse methods including LTH, gradual magnitude pruning (GMP), and pruning at initialization (PI). Our results show that, with a sufficient and reliable parameter exploration, DST consistently outperforms those overparameterization-based sparse methods.
% We apply ITOP to the main sparse training method, DST, and compare it against the overparameterization-based sparse methods including LTH, gradual magnitude pruning (GMP), and pruning at initialization (PI). Our results show that, with a sufficient and reliable parameter exploration, DST consistently outperforms those overparameterization-based sparse methods and ITOP is of vital importance for sparse training. 

\section{Related Work}
\subsection{Dense Over-Parameterization} 
Sparsity-inducing techniques that depend on dense over-parameterization (dense-to-sparse training) have been extensively studied. We divide them into three categories according to their degrees of dependence on the dense over-parameterization.

\textbf{Full dense over-parameterization.} Techniques sought to inherit weights from a fully pre-trained dense model have a long history and were first introduced by \citet{janowsky1989pruning} and \citet{mozer1989using}, autonomously evolving as the iterative pruning and retaining method. The basic idea of iterative pruning and retaining involves a three-step process: (1) fully pre-training a dense model until converged, (2) pruning the weights or the neurons that have the lowest influence on the performance, and (3) re-training the pruned model to further improve the performance. The pruning and retraining cycle is required at least once \cite{liu2019rethinking}, and usually many times \citep{han2015deep,guo2016dynamic,frankle2018the}. The criteria used for pruning includes but are not limited to magnitude \citep{mozer1989using,han2015deep,guo2016dynamic}, Hessian \citep{lecun1990optimal,hassibi1993second}, mutual information \citep{dai2018compressing}, Taylor expansion \citep{molchanov2016pruning,molchanov2019importance}. Except for pruning, other techniques including variational dropout \citep{molchanov2017variational}, targeted dropout \citep{gomez2019learning}, reinforcement learning \citep{lin2017runtime} also yield a sparse model from a pre-trained dense model.

% The training costs and memory requirements of the aforementioned methods are at least the same as fully training a dense network, sometimes much more. 

\textbf{Partial dense over-parameterization.} Another class of methods start from a dense network and continuously sparsify the model during training.  Gradual magnitude pruning (GMP) \citep{narang2017exploring,zhu2017prune,gale2019state} was proposed to reduce the number of pruning-and-retaining rounds by pruning the dense network to the target sparsity gradually over the course of training. There are some examples \citet{louizos2017learning} and \citet{wen2016learning} that utilize $L_0$ and $L_1$ regularization to gradually learn the sparsity by explicitly penalizing parameters for being different from zero, respectively. Recently, \citet{srinivas2017training,LIU2020Dynamic,savarese2019winning,xiao2019autoprune,kusupati2020soft,zhou2021learning} moved further by introducing trainable masks to learn the desirable sparse connectivity during training. Since these techniques start from a dense model, the training cost is smaller than training a dense network, depending on the stage at which the final sparse models are learned.

\textbf{One-Shot dense over-parameterization.} Very recently, works on pruning at initialization (PI) \citep{lee2018snip,Lee2020A,Wang2020Picking,tanaka2020pruning,jorge2021progressive} have emerged to obtain trainable sparse neural networks before the main training process based on some salience criteria. These methods fall into the category of dense over-parameterization mainly because the dense model is required to train for at least one iteration to obtain those trainable sparse networks.

\subsection{In-Time Over-Parameterization} 
\textbf{Dynamic Sparse Training.} Evolving in parallel with LTH, DST is a growing class of methods to train sparse networks from scratch with a fixed parameter count throughout training (sparse-to-sparse training).  This paradigm starts from a (random) sparse neural network and allows the sparse connectivity to evolve dynamically during training. It has been first introduced in \citet{setphdthesis2017} and became well-established in \citet{mocanu2018scalable} by proposing the Sparse Evolutionary Training (SET) algorithm which achieves better performance than static sparse neural networks. In addition to the proper classification performance, it also helps to detect important input features \citep{atashgahi2020quick}. \citet{bellec2018deep} proposed Deep Rewiring to train sparse neural networks with a strict connectivity constraint by sampling sparse configurations and weights from a posterior distribution. Follow-up works further introduced weight redistribution \citep{mostafa2019parameter,dettmers2019sparse,liu2021selfish}, gradient-based weight growth \citep{dettmers2019sparse,evci2019rigging}, and extra weights update in the backward pass \citep{raihan2020sparse,jayakumar2020top} to improve the sparse training performance. By relaxing the constraint of the fixed parameter count, \citet{dai2019nest,dai2018grow} proposed a grow-and-prune strategy based on gradient-based growth and magnitude-based pruning to yield an accurate, yet very compact sparse network.  %Nevertheless, due to the inherent limitations of deep learning software and hardware libraries, all of the above works simulate sparsity using a binary mask over the dense weights. 
More recently, \citet{onemillionneurons} illustrated for the first time the true potential of using dynamic sparse training. By developing an independent framework, they can train truly sparse neural networks without masks with over one million neurons on a typical laptop.
% Cross-layer weights reallocation was introduced in Dynamic parameter reallocation (DSR)  by \citet{mostafa2019parameter}. DSR is able to allocate more parameters to layers where the training loss drops more efficiently. Sparse Networks from Scratch (SNFS) \citep{dettmers2019sparse} leverages the momentum of zero-valued weights to guide the optimization of sparse connectivity. Recently, Rigged Lottery (RigL) \citep{evci2019rigging} chooses connections with high gradient magnitude to perform periodic updates on the sparse connectivity and can match the performance of pruning methods given 5$\times$ training time. Nevertheless, due to the inherent limitations of deep learning software and hardware libraries, all of the above works simulate sparsity using a binary mask over weights. More recently, \cite{onemillionneurons} has illustrated for the first time the true potential of using dynamic sparse training, by developing an independent framework from scratch to train truly sparse neural networks and being able to train sparse MLPs with over one million neurons on a typical laptop. 

\textbf{Understanding Dynamic Sparse Training.} Concurrently, some works attempt to understand Dynamic Sparse Training. \citet{liu2020topological} found that DST gradually optimizes the initial sparse topology towards a completely different one. Although there exist many low-loss sparse solutions that can achieve similar loss, they are very different in the topological space. \citet{evci2020gradient} found that sparse neural networks that are initialized by a dense initialization e.g., \citet{he2015delving}, suffer from a poor gradient flow, whereas DST can improve the gradient flow during training significantly. Although promising, the capability of sparse training has not been fully explored and the mechanism underlying DST is not clear yet. Questions like: \textit{Why Dynamic Sparse Training can improve the performance of sparse training? How Dynamic Sparse Training can enable sparse neural network models to match - and even to outperform - their dense counterparts?} are required to be answered.

\section{In-Time Over-Parameterization}
% By observing that Dynamic Sparse Training significantly improves the performance of Static Sparse Training simply by optimizing the sparse connectivity together with the model parameters, 

In this section, we describe in detail In-Time Over-Parameterization, a concept that we proposed to be an alternative way to train deep neural networks without the expensive over-parameterization. We refer In-Time Over-Parameterization as a variant of dense over-parameterization, which can be achieved by encouraging a continuous parameter exploration across the training time. Note that different from the over-parameterization of dense models which refers to the spatial dimensionality of the parameter space, In-Time Over-Parameterization refers to the overall dimensionality explored in the space-time manifold.

\subsection{In-Time Over-Parameterization Hypothesis}
Based on In-Time Over-Parameterization, we propose the following hypothesis to understand Dynamic Sparse Training:

\textbf{Hypothesis.} \textit{The benefits of Dynamic Sparse Training come from its ability to consider across time all possible parameters when searching for the optimal sparse neural network connectivity}. Concretely, this hypothesis can be divided into three main pillars which can explain the performance of DST: 
\begin{enumerate}
  \item Dynamic Sparse Training can significantly improve the performance of sparse training mainly due to the parameter exploration across the training time.
  \item The performance of Dynamic Sparse Training is highly related to the total number of the reliably explored parameters throughout training. The reliably explored parameters refer to those newly-explored (newly-grown) weights that have been updated for long enough to exceed the pruning threshold.
  \item As long as there are sufficient parameters that have been reliably explored, sparse neural network models trained by Dynamic Sparse Training can match or even outperform their dense counterparts by a large margin, even at extremely high sparsity levels.
\end{enumerate}

We name our hypothesis as In-Time Over-Parameterization hypothesis for convenience. 

Formally, given a dataset containing $N$ samples $\mathbf{D} = \{(x_i, y_i) \}_{i=1}^N$ and a dense network $f(x; \theta)$ parameterized by $\theta$. We train the dense network to minimize the loss function $\sum_{i=1}^{N} L(f(x_i;\theta),y_i)$. When optimizing with a certain optimizer, $f(x; \theta)$ reaches a minimum validation loss function $l$ with a test accuracy $a$. Differently, sparse training starts with a sparse neural network $f(x; \theta_s)$ parameterized by a fraction of parameters $\theta_s$. The basic mechanism of Dynamic Sparse Training is to train the sparse neural network $f(x; \theta_s)$ to minimize the loss $\sum_{i=1}^{N} L(f(x_i;\theta_s),y_i)$ while periodically update the sparse connectivity $\theta_s$ every $\Delta T$ iterations based on some criteria. $f(x; \theta_s^u)$ reaches a minimum validation loss $l'$ at sparse connectivity update $u$ with a test accuracy $a'$, where $\theta_s^u$ is the sparse connectivity parameters obtained at the iteration $u$. Let us denote $R_s$ as the ratio of the total number of reliably explored parameters during training to the total number of parameters, or simply In-Time Over-Parameterization rate, computed as $R_s=\frac{\|\theta_s^1 \cup \theta_s^2 \cup...\cup \theta_s^u \|_0}{\|\theta\|_0}$, where $\|\cdot\|_0$ is the $\ell_0$-norm.

Our hypothesis states that when $\Delta T \geq T_0$, $\exists R_0$ as long as $R_s \geq R_0$, for which $a' \geq a$ (commensurate accuracy) and $\|\theta_s^u\|_0 \ll \|\theta\|_0$ (fewer parameters in the final sparse model), where $T_0$ is the minimum threshold of update interval to guarantee the reliable parameter exploration, and $R_0$ is the threshold of In-Time Over-Parameterization rate where DST can match the performance of the dense model.

Similar to dense training, there are many factors affecting the performance of dynamic sparse training, learning rate, batch size, regularization, optimizers, sparsity distribution, etc. In this paper, we limit our study to parameter exploration, since it is the fundamental difference between dynamic sparse training and static sparse training. By “reliable”, we mainly focus on the newly-activated weights that are updated for a long time (guaranteed by $\Delta T \geq T_0$) so that they are not pruned in the next update iteration. We believe this is a good starting point, since even in this simple setting, our community does not have a satisfactory answer.

\begin{figure*}[t]
\vspace{-0.1em}
        \centering
        \includegraphics[width=\textwidth]{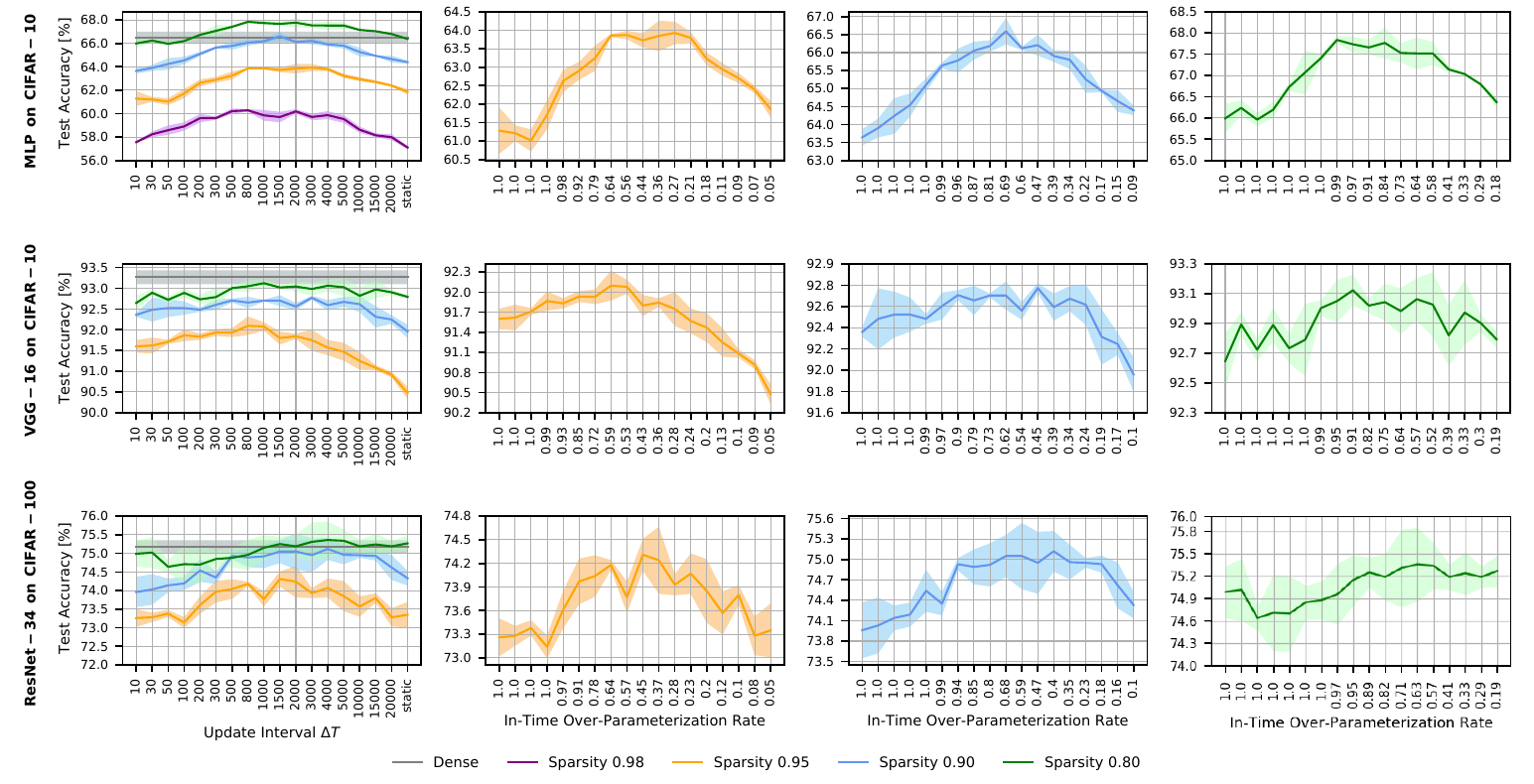}
        \vspace{-0.5cm}
        \caption[ The average and standard deviation of critical parameters ]
        {\small Effect of In-Time Over-Parameterization on sparse training MLPs (\textbf{top}), VGG-16 (\textbf{middle}), and ResNet-34 (\textbf{bottom}) with a typical training time. All sparse models are trained with SET. Each line is averaged from three different runs. ``Static'' refers to the static sparse training without parameter exploration.  }  
        \label{fig:typical_time}
        \vspace{-0.1em}
    \end{figure*}
    
\subsection{Hypothesis Evaluation}
\label{sec:hy_eva}
In this section, we work through the In-Time Over-Parameterization hypothesis and study the effect of In-Time Over-Parameterization on the performance of DST. We choose Sparse Evolutionary Training (SET) as our DST method as SET activates new weights in a random fashion which naturally considers all possible parameters to explore. It also helps to avoid the dense over-parameterization bias introduced by the gradient-based methods e.g., The Rigged Lottery (RigL) \citep{evci2019rigging} and Sparse Networks from Scratch (SNFS) \citep{dettmers2019sparse}, as the latter utilize dense gradients in the backward pass to explore new weights. To work through the proposed hypothesis, we conduct a set of step-wise fashion experiments with image classification. We study Multi-layer Perceptron (MLP) on CIFAR-10, VGG-16 on CIFAR-10, ResNet-34 on CIFAR-100, and ResNet-50 on ImageNet. We use PyTorch as our library. All results are averaged from three different runs and reported with the mean and standard deviation. See Appendix \ref{Imp_settings} for the experimental details.

\subsubsection{Typical Training Time}
Our first evaluation of the In-Time Over-Parameterization hypothesis is to see what happens when different over-parameterization rates $R_s$ are reached during training within a typical training time (200 or 250 epochs). A direct way to control $R_s$ is to vary $\Delta T$, a hyperparameter that determines the update interval of sparse connectivities (the number of iterations between two sparse connectivity updates). We train MLP, VGG-16, and ResNet-34 with various $\Delta T$ and report the test accuracy. 

\textbf{Expected results.} Gradually decreasing $\Delta T$ will explore more parameters, and thus lead to increasingly higher test accuracy. However, when $\Delta T$ gets smaller than the reliable exploration threshold $T_0$, the test accuracy will start to decrease since the new weights can not receive enough updates to exceed the pruning threshold.

\textbf{Experimental results.} For a better overview, we plot the performance achieved at different sparsities together in the leftmost column of Figure \ref{fig:typical_time}. To understand better the relationship between $R_s$ and test accuracy, we report the final $R_s$ associated with $\Delta T$ separately in the rest columns of Figure \ref{fig:typical_time}.

Overall, a similar pattern can be found existing in all lines. Starting from the static sparse training, sparse training consistently benefits from the increased $R_s$ as $\Delta T$ decreases. However, the test accuracy starts to drop rapidly after it reaches a peak value, especially at high sparsities (yellow and blue lines). For example, even if MLPs and ResNet-34 eventually reach a 100\% exploration rate with extremely small $\Delta T$ values (e.g., 10, 30), their performance is much worse than the static sparse training. This behavior is perfectly in line with our hypothesis. While small $\Delta T$ encourages sparse models to maximally explore the search space spanned over the dense model, the benefits provided by In-Time Over-Parameterization is heavily limited by the unreliable parameter exploration. Interestingly, the negative effect of the unreliable exploration on lower sparsities (green lines) is less than the one on high sparsities (yellow lines). We regard this as trivial sparsities \cite{frankle2020linear} as the remaining models are still over-parameterized to fit the data. 

% \textcolor{red}{In other words}, smaller $\Delta T$ encourages the sparse model to maximally explore the search space in terms of sparse connectivities spanned over the dense model.

\subsubsection{Extended Training Time}
Until now, we have already learned the trade-off between test accuracy and $R_s$ for the typical training time. A direct approach to alleviating this trade-off is to extend the training time while using large $\Delta T$. We train MLP, VGG-16, and ResNet-34 for an extended training time with a large $\Delta T$. We safely choose $\Delta T$ as 1500 for MLPs, 2000 for VGG-16, and 1000 for ResNet-34 according to the trade-off shown in Figure \ref{fig:typical_time}. In addition to the training time, the anchor points of the learning rate schedule are also scaled by the same factor. 
\begin{figure*}[ht]
\vspace{-0.1em}
        \centering
        \includegraphics[width=\textwidth]{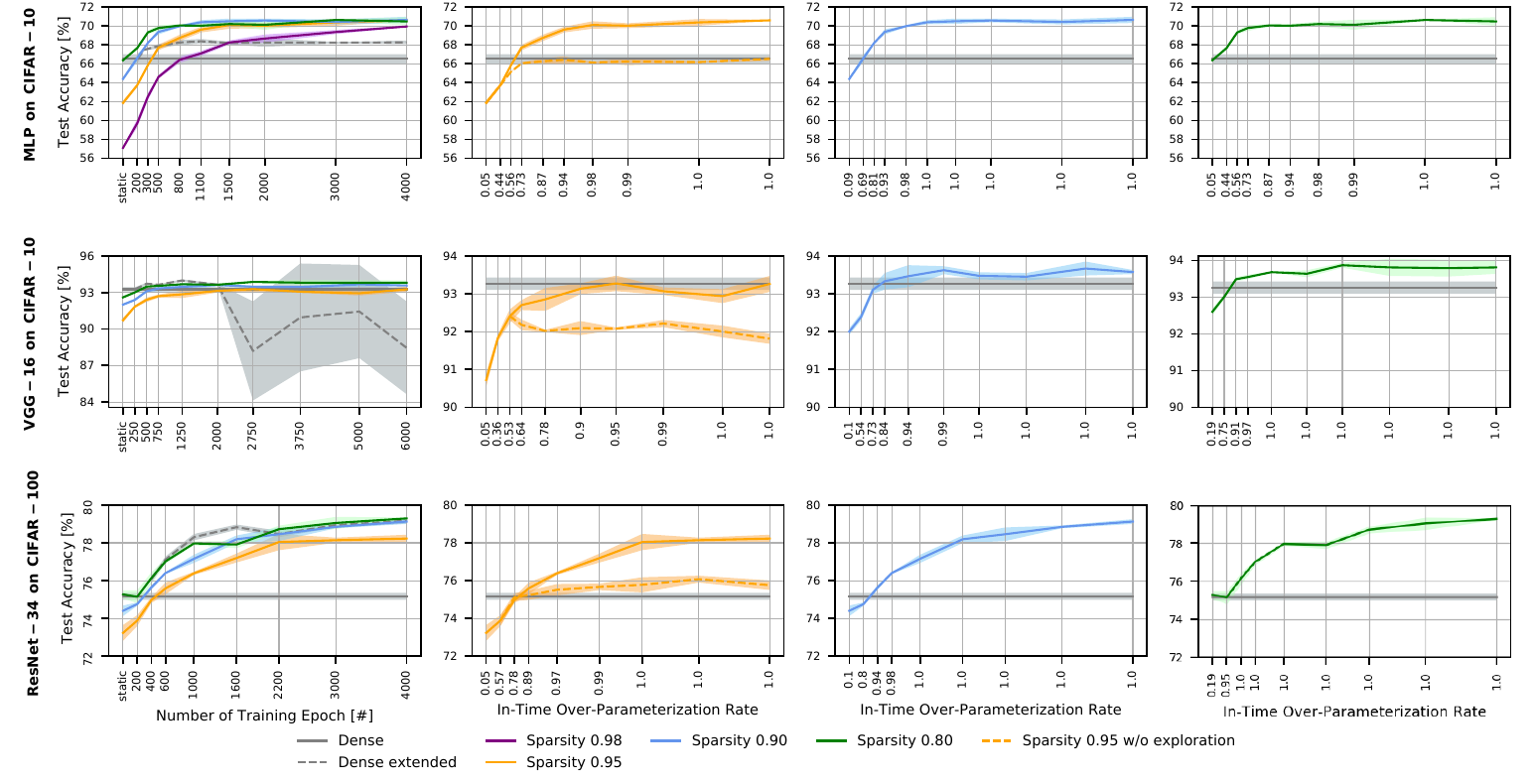}
        \vspace{-0.5cm}
        \caption[ The average and standard deviation of critical parameters ]
        {\small Effect of In-Time Over-Parameterization on sparse training MLPs (\textbf{top}), VGG-16 (\textbf{middle}) and ResNet-34 (\textbf{bottom}) with an extended training time. All sparse models are trained with SET. Each line is averaged from three different runs. ``Static'' refers to the static sparse training without parameter exploration. ``Dense extended'' refers to training a dense model for an extended time. ``Sparsity 0.95 w/o exploration'' means we train the model for the same extended time but stop exploring parameters after a typical training time (200 or 250 epochs).}  
        \label{fig:Various_T_extime}
        \vspace{-0.1em}
    \end{figure*}
    
\textbf{Expected results.} In this setting, we expect that, in addition to the benefits brought by the extended training time, sparse training would benefit significantly from the increased $R_s$.

\textbf{Experimental results.} The results are shown in Figure \ref{fig:Various_T_extime}. Static sparse training without parameter exploration consistently achieves the lowest accuracy. However, all models at different sparsities substantially benefit from an extended training time accompanied by an increased $R_s$. In other words, reliably exploring the parameter space in time continuously improves the expressibility of sparse training. Importantly, after matching the performance of the dense baseline (black line), the performance of sparse training continues to improve, yielding a notable improvement over the dense baseline. Furthermore, the models with lower sparsities require less time to match their full accuracy plateau than higher sparsities; the cause appears to be that models with lower sparsity can explore more parameters in the same training time. 

To show that the performance gains are not only caused by the longer training time, we make a controlled experiment by stopping the parameter exploration immediately after the typical training time (the sparse connectivity remains fixed after a typical training time), shown as the orange dashed lines. As we can see, even though improved, the accuracy is much lower than the accuracy achieved by In-Time Over-Parameterization.

We also report the performance of dense models with an extended training time as the dashed black lines. Training a dense model with an extended time leads to either inferior (MLPs and VGG-16), or equal solutions (ResNet-34). Different from the dense over-parameterization where overfitting usually occurs when the model has been overtrained for long enough, the test accuracy of dynamic sparse training is continuously increasing as $R_s$ increases until a plateau is reached with a full In-Time Over-Parameterization. This observation highlights the advantage of In-Time Over-Parameterization to prevent overfitting over the dense over-parameterization.

\section{Effect of Hyperparameter Choices}

\subsection{Effect of Weight Growth methods on ITOP}
\begin{figure*}[ht]
\vspace{-0.1em}
    \centering
    \includegraphics[width=\textwidth]{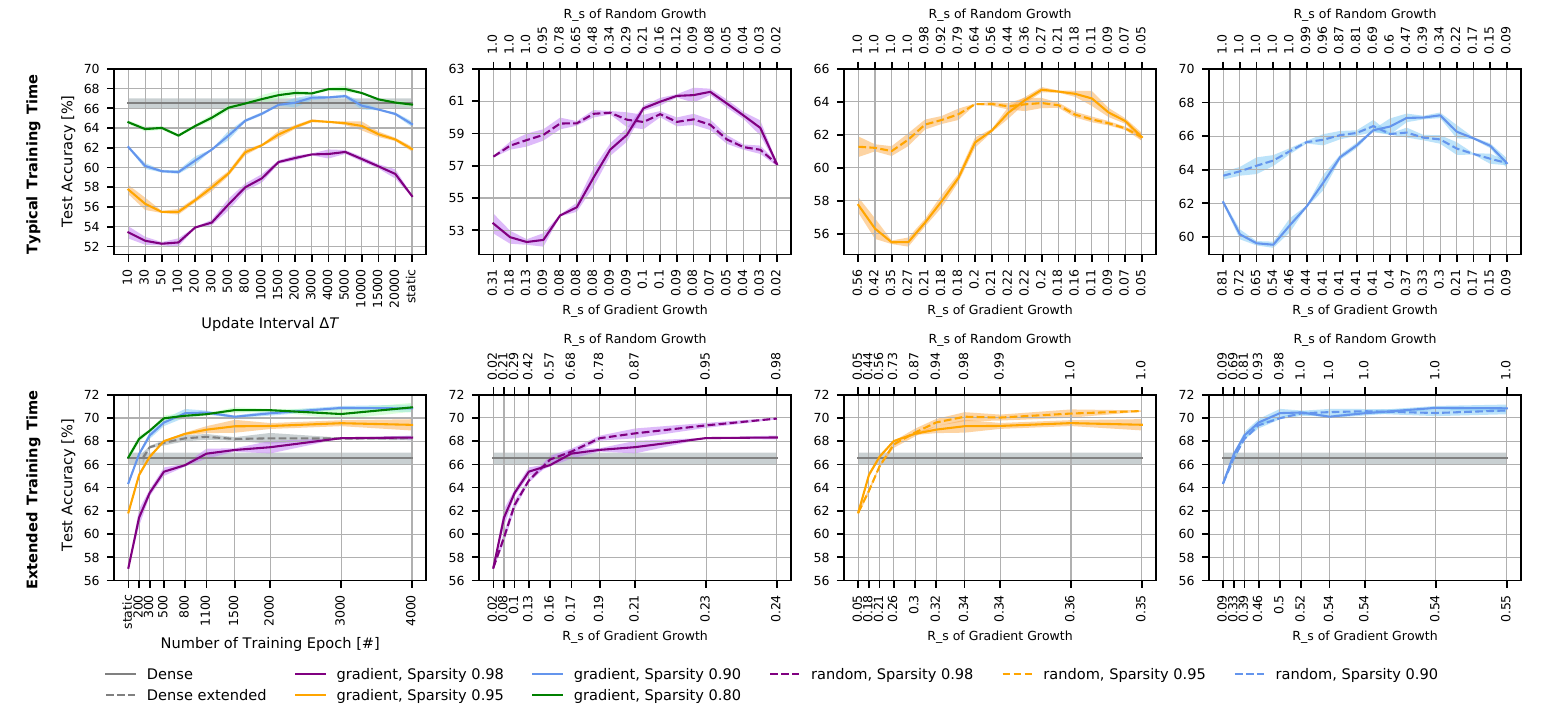}
    \vspace{-0.8cm}
    \caption
        {\small Comparisons between RigL and SET with MLP on CIFAR-10. We vary the update interval $\Delta T$ for the typical training time setting, and keep it fixed for the extended training time setting (1500 for SET and 4000 for RigL).}  
    \label{Fig:rigl}
    \vspace{-0.3em}
\end{figure*}

We next investigate the effect of gradient-based weight growth (used in RigL and SNFS) and random-based weight growth (used in SET) on In-Time Over-Parameterization. Since gradient-based methods have access to the dense over-parameterization in the backward pass (occasionally using dense gradients to activate new weights), we hypothesize that they can reach a converged accuracy without a high $R_s$. We make a comparison between RigL and SET for both the typical training and the extended training in Figure \ref{Fig:rigl}. We study them on MLPs where the model size is relatively small so that we can easily achieve a full In-Time Over-Parameterization and have a better understanding of these two methods.

\textbf{Typical Training Time.} 
It is clear that RigL also heavily suffers from the unreliable exploration. As $\Delta T$ decreases, the test accuracy of RigL presents a trend of rising, falling, and rising again. Compared with the random-based growth, RigL receives larger gains from the reliable parameter exploration and also a larger forfeit from the unreliable exploration. These differences are potentially due to that RigL grows new weights with high gradient magnitude, which leads to a faster loss decrease when the exploration is faithful, but also requires a higher $\Delta T$ to guarantee a faithful exploration as the weight with large gradients is likely to end up with high magnitude, resulting in a large pruning threshold.

\textbf{Extended Training Time.} 
For RigL, we choose $\Delta T = 4000$ to ensure the reliable exploration (the performance of RigL with a smaller $\Delta T = 1500$ is much worse as shown in Appendix \ref{app:rigl_1500}). We can see that RigL also significantly benefits from an increased $R_s$. Surprisingly, although RigL achieves higher accuracy than the SET with a limited training time, it ends up with lower accuracy than SET with a sufficient training time. From the perspective of $R_s$, we can see that the $R_s$ of RigL is much smaller than SET, indicating that gradient weight growth drives the sparse connectivity into some similar structures and in turn limits its expressibility. On the contrary, random growth naturally considers the whole search space to explore parameters and has a larger possibility of finding better local optima.  Similar results are also reported for sparse Recurrent Neural Networks (RNNs) in \citet{liu2021selfish}. However, similar results are not shared with large-scale architectures on large datasets. For instance, RigL achieves better performance than SET with ResNet-50 on ImageNet. This result is reasonable since the dense gradients help RigL easily find the most promising weights at each sparse connectivity update. In contrast, it would take a much longer time (high $R_s$) for SET (random weight growth) to discover these promising weights within large-scale architectures, especially at high sparsities.

\vspace{-0.3em}
\subsection{Effect of Batch Size on ITOP}
\vspace{-0.3em}
\label{sec:impro_dst}

\begin{figure*}[h]
\vspace{-0.5em}
    \centering
    \includegraphics[width=\textwidth]{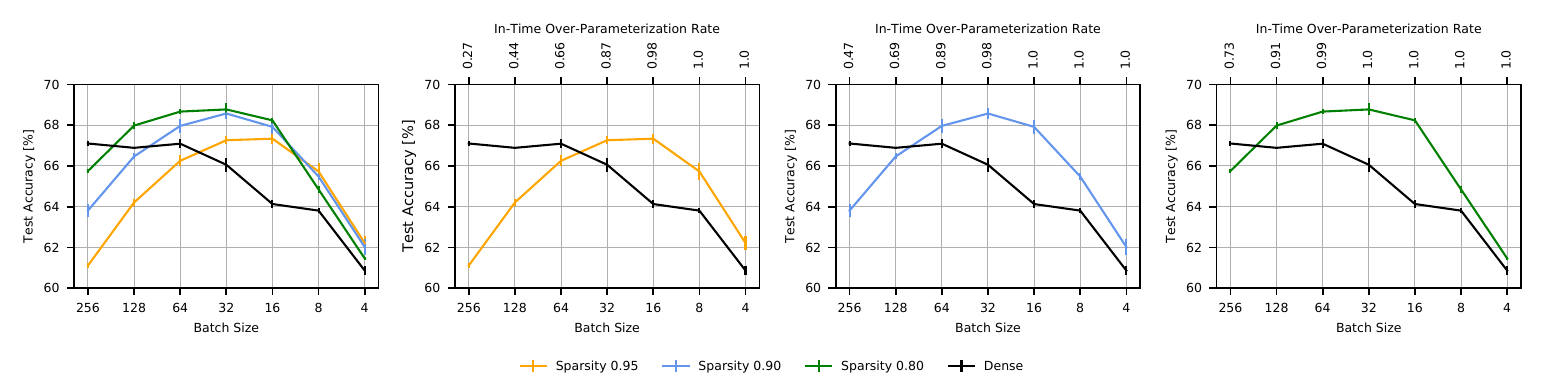}
    \vspace{-0.7cm}
    \caption
    {\small Test accuracy of SET with various batch sizes. The update interval $\Delta T$ is set as 1500.}  
    \label{Fig:set_various_bs}
    \vspace{-0.3em}
\end{figure*}

Intuitively, our hypothesis uncovers ways to improve the existing DST methods within a limited training time. A direct way to reliably explore more parameters within a typical training time is to train with a small batch size. Using a smaller batch size equally means having more updates, and therefore leads to a higher $R_s$. We simply demonstrate the effectiveness of this conjecture on SET with $\Delta T=1500$ in Figure \ref{Fig:set_various_bs} (see Appendix \ref{app:rigl_bs} for RigL). With a large batch size, the parameter exploration is insufficient to achieve a high In-Time Over-Parameterization rate $R_s$, and the test accuracy is subsequently much lower than the dense model. As we expected, the reduction in batch size consistently increases $R_s$ as well as the test accuracy, until the batch size gets smaller than 16. However, the performance of the dense model remarkably decreases as the batch size decreases. More interestingly, when the batch size is smaller than 16, the performance of sparse models flips and the sparsest model starts to achieve the highest accuracy. The performance drop is likely caused by the increased “noise scale” of SGD where extremely small batch sizes lead to large noise scale and large accuracy drop (\citet{smith2017don}).

\begin{table*}[h]
\scriptsize
\vspace{-0.3em}
\centering
\caption{Performance of sparse ResNet-18 on CIFAR-10 with various pruning rates. The results are run three times and reported with $(\textrm{mean}\pm\textrm{std}, R_s)$. The highest test accuracies are marked in bold.}
\label{tab:pruning_rate}
\resizebox{1.0\textwidth}{!}{
\begin{tabular}{c c  c| c c ccc}
\toprule
Sparsity &  Method &  P & $\Delta T=15000$ & $\Delta T=10000 $  & $\Delta T=8000 $ & $\Delta T=5000 $ & $\Delta T=3000 $  
\\ 
\midrule
\multirow{5}{*}{0.95} & SET & 0.9 & (93.53 $\pm$ 0.02, 0.113) & (93.57 $\pm$ 0.05, 0.150) & (93.44 $\pm$ 0.12, 0.176) & (93.70 $\pm$ 0.19, 0.247) & (93.78 $\pm$ 0.09, 0.353)
\\
& SET & 0.7 & ({\bf 93.54} $\pm$ 0.09, 0.100) & (93.54 $\pm$ 0.18, 0.130) & (93.76 $\pm$ 0.07, 0.151) & (93.91 $\pm$ 0.17, 0.210) & (93.63 $\pm$ 0.01, 0.300)
\\
& SET & 0.5 &  (93.51 $\pm$ 0.01, 0.086) & ({\bf 93.77} $\pm$ 0.21, 0.109) & ({\bf 93.84} $\pm$ 0.10, 0.125) & ({\bf 93.93} $\pm$ 0.09, 0.170) & ({\bf 93.94} $\pm$ 0.08, 0.241) 
\\ 
& SET & 0.3 & (93.28 $\pm$ 0.01, 0.071) & (93.66 $\pm$ 0.05, 0.086) & (93.80 $\pm$ 0.01, 0.096) & (93.75 $\pm$ 0.18, 0.126) & (93.86 $\pm$ 0.10, 0.174)
\\ 
& SET & 0.1 & (93.24 $\pm$ 0.02, 0.056) & (93.35 $\pm$ 0.18, 0.061) & (93.29 $\pm$ 0.06, 0.065) & (93.50 $\pm$ 0.03, 0.076) & (93.34 $\pm$ 0.03, 0.096) 
\\
\bottomrule
\end{tabular}}
\vspace{-0.3em}
\end{table*}

% \vspace{-0.3em}
\subsection{Effect of Pruning Rate on ITOP}
% \vspace{-0.3em}
The initial pruning rate (denoted as $P$) of parameter exploration also affects the overall number of parameters visited during training. Relatively large pruning rates encourage a large range of exploration, resulting in higher accuracy, whereas a too-large pruning rate hurts the model capacity as it prunes too many parameters. We confirm this with ResNet-18 on CIFAR-10 trained with various initial pruning rates $P \in [0.1, 0.3, 0.5, 0.7, 0.9]$ as shown in Table~\ref{tab:pruning_rate}. The similar pattern as we expected is shared across all update intervals.

\subsection{Boosting the Performance of DST}
Based on the above-mentioned insights, we demonstrate the state-of-the-art sparse training performance with ResNet-50 on ImageNet. More precisely, we choose an update interval $\Delta T $ of 4000, a batch size of 64, and an initial pruning rate of 0.5 so that we can achieve a high $R_s$ within a typical training time. We briefly name the improved method as RigL-ITOP. Please see Appendix \ref{App_riglITOP} for the implementation details. Table \ref{tab:ImageNet} shows that without any advanced techniques, our method boosts the accuracy of RigL over the overparameterization-based method (GMP and Lottery Ticket Rewinding (LTR) \citep{frankle2020linear}). More importantly, our method requires only $2\times$ training time to match the performance of dense ResNet-50 at 80\% sparsity, far less than RigL ($5\times$ training time) \citep{evci2019rigging}. 

Instead of using small batch size, another trick to encourage parameter exploration is sampling from the non-activated weights first when growing new weights. We demonstrate the effectiveness of this idea in Appendix~\ref{app:samping}.

\begin{table*}[ht]
\scriptsize
\vspace{-0.3em}
\centering
\caption{Performance of sparse ResNet-50 on ImageNet dataset with a typical training time. All results of other methods are obtained from \citet{evci2019rigging} except LTR which is the late-rewinding LTH version obtained from \citet{evci2020gradient}. RigL-ITOP$_{2\times}$ is obtained by extending the training time by 2 times.}
\label{tab:ImageNet}
\resizebox{1.0\textwidth}{!}{
\begin{tabular}{c |c  c c c| cccc}
\toprule
Methods & Top-1 Acc & $R_s$ & Training & Test & Top-1 Acc & $R_s$ & Training & Test \\
& & & FLOPs & FLOPs & & & FLOPs & FLOPs \\
\midrule
Dense & $76.8 \pm 0.09$ & 1.00 & 1$\times$ (3.2e18) & 1$\times$ (8.2e9) & $76.8 \pm 0.09$ & 1.00 & 1$\times$ (3.2e18) & 1$\times$ (8.2e9)  \\
\midrule
& \multicolumn{4}{c|}{sparsity=0.9} & \multicolumn{4}{c}{sparsity=0.8} \\
\midrule
Static  & $67.7 \pm 0.12$ & 0.10 & 0.24$\times$ & 0.24$\times$ & $72.1 \pm 0.04$ & 0.20  & 0.42$\times$ & 0.42$\times$\\
SET  & $69.6 \pm 0.23$ & - & 0.10$\times$ & 0.10$\times$ & $72.9 \pm 0.39$  & - & 0.23$\times$ & 0.23$\times$ \\
SNFS & $72.9 \pm 0.06$ & - & 0.50$\times$ & 0.24$\times$ & $75.2 \pm 0.11$  & - & 0.61$\times$ & 0.42$\times$ \\
RigL & $73.0 \pm 0.04$ & - & 0.25$\times$ & 0.24$\times$  & $75.1 \pm 0.05$ & - & 0.42$\times$ & 0.42$\times$  \\

GMP & 73.9 & - & 0.56$\times$ & 0.23$\times$ & $75.6$ & -  & 0.51$\times$ & 0.10$\times$ 
\\
LTR & - & - & - & - & $75.75 \pm 0.12$ & - & - & -
% \\
% SET-ITOP & 65.2 & 1.00 & 0.10$\times$ & 0.10$\times$ & 71.9 & 1.00 & 0.23$\times$ & 0.23$\times$ 
\\
RigL-ITOP & ${73.82 \pm 0.08}$ & 0.83 & 0.25$\times$ & 0.24$\times$ & ${75.84 \pm 0.05}$ & 0.93 &0.42$\times$ & 0.42$\times$  
\\
RigL-ITOP$_{2\times}$ & $\mathbf{75.50 \pm 0.09}$ & 0.89 &
0.50$\times$ & 0.24$\times$ & $\mathbf{76.91 \pm 0.07}$ & 0.97 & 0.84$\times$ & 0.42$\times$ 
\\
% (Ours) & & &  & \\
\bottomrule
\end{tabular}}
\vspace{-0.3em}
\end{table*}

\section{The Versatility of ITOP}
\label{sec:comparison}

Although we mainly focus on understanding DST from the ITOP point of view, ITOP can be potentially generalized to other sparsity-inducing categories. Here, we demonstrate its versatility by applying ITOP to two recently popular methods, LTH and PI. We choose SNIP~\citep{lee2018snip} as the PI
method, as it consistently performs well among different methods for pruning at initialization as shown by~\citet{frankle2020pruning}. Compared with ITOP, LTH and SNIP are two overparameterization-based methods designed for a better initial subnetwork but without consulting any information yielded during training.  We choose magnitude-based weight pruning and random-based weight growth for SNIP and LTH to achieve ITOP and name the corresponding methods as SNIP-SET-ITOP and LTH-SET-ITOP. To make a fair comparison between different pruning criteria, we use global and one-shot pruning for both SNIP and LTH. We train all models for 200 epochs and report the best test accuracy in Figure~\ref{Fig:comparisons}. See Appendix \ref{Rep_other_methods} for the experimental details.

\begin{figure}[h]
\vspace{-0.2em}
    \centering
    \includegraphics[width=0.48\textwidth]{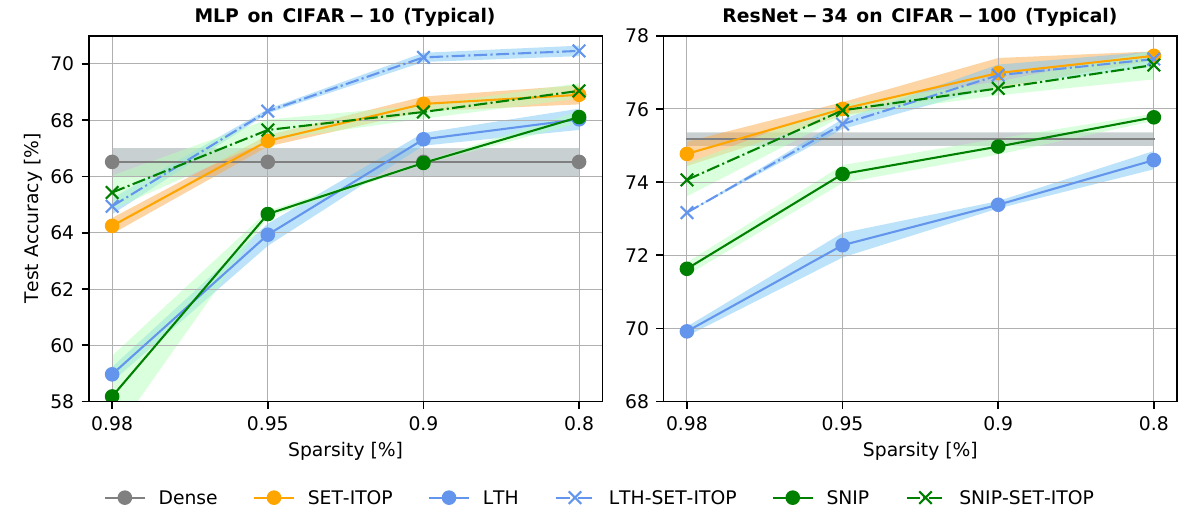}
    \caption
    {\small Effect of In-Time Over-Parameterization on SNIP and LTH. ``Typical'' means training a model for a typical time of 200 epochs.}  
    \label{Fig:comparisons}
    \vspace{-0.1em}
\end{figure}
With a high In-Time Over-Parameterization rate, SET-ITOP consistently outperforms the overparameterization-based methods as well as the dense training, by a large margin. For instance, SET-ITOP can easily match the performance of the corresponding dense models with at most 5\% parameters. More importantly, SET-ITOP has dominant performance at the extreme sparsity (98\%) over LTH and SNIP, indicating the potential of In-Time Over-Parameterization to address the poor expressibility problem of the extremely sparse neural networks.

It is maybe more interesting that In-Time Over-Parameterization brings large benefits to SNIP and LTH as well. While LTH and SNIP fall short of SET-ITOP, SNIP-SET-ITOP and LTH-SET-ITOP can match or even exceed the performance of SET-ITOP with both MLP and ResNet-34. This observation confirms that ITOP is a foundational concept and can potentially improve any existing sparse training methods.

% SNIP generally achieves better performance than static sparse training for all settings with only one iteration of dense training. While LTH performs well with MLP, it is the worst-performing method with ResNet-34 even with the information inherited from a fully converged dense model. This observation makes a connection with the findings in \citet{frankle2020linear} which demonstrates that, in large-scale settings (e.g., ResNet-50 on ImageNet), subnetworks uncovered by iterative magnitude pruning only can train to the same accuracy as the full network after the full network has been trained for some number of epochs rather than at initialization. Moreover, the performance of GMP with ResNet-34 significantly benefits from an extended training time.

% only found by iterative magnitude pruning early in training rather than at initialization 
% that only considering magnitude as the pruning criterion for sparse training is not optimal. Techniques that utilize the combination of weight and gradient can provide better guidance for the after-pruning training states. 

\vspace{-0.1em}
\section{Generalization Improvement of ITOP}
\vspace{-0.1em}
\label{generalization}

\begin{figure}[h]
\vspace{-0.1em}
        \centering
        \includegraphics[width=0.48\textwidth]{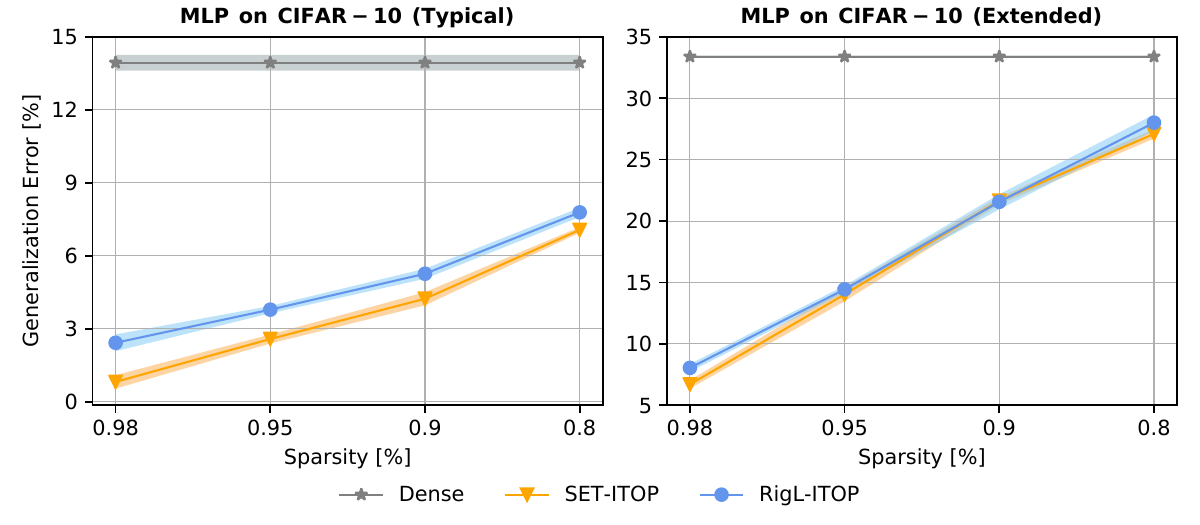}
        % \vspace{-0.6cm}
        \caption[ The average and standard deviation of critical parameters ]
        {\small Generalization errors of SET-ITOP, RigL-ITOP, and the dense models.}
        \label{fig:generalization}
\vspace{-0.1em}
\end{figure}

We further observe the ability of In-Time Over-Parameterization to improve generalization. Figure \ref{fig:generalization} shows that the generalization error (the difference between the training accuracy and the test accuracy) of In-Time Over-Parameterization (SET-ITOP and RigL-ITOP) and the dense over-parameterization with MLPs on CIFAR-10. It is clear to see that models with the In-Time Over-Parameterization property generalize much better than the corresponding dense models. The generalization error gradually increases as the model gets denser. Together with the results in Figure \ref{Fig:comparisons}, we can see that the reductions in sparsity lead to better classification performance but worse generalization.

\vspace{-0.1em}
\section{Conclusion and Future Work}
\vspace{-0.1em}
In this paper, we propose In-Time Over-Parameterization, a variant of dense over-parameterization in the space-time manifold, to be an alternative way to train deep neural networks without the prohibitive dense over-parameterized dependency. We demonstrate the ability of In-Time Over-Parameterization (1) to improve the expressibility of sparse training, (2) to accelerate both training and inference, (3) to understand the underlying mechanism of DST, (4) to prevent overfitting and improve generalization. In addition, we empirically found that, with a sufficient and reliable parameter exploration, randomly-initialized sparse models consistently achieve better performance over those specially-initialized static sparse models.  Our paper suggests that it is more effective and efficient to allocate the limited resources to explore more the sparse connectivity space, rather than allocating all resources to find a good sparse initialization.

Our paper discovers the importance of parameter exploration for sparse training. Even though we adjust the hyperparameters of RigL and reach the state of art sparse training performance, the usage of small batch size slows down the training speed of modern architectures. It is interesting to pursue a high In-Time Over-Parameterization rate with large batch size under a typical training time. Moreover, we believe that ITOP has potentials to help people to interpret the networks' decisions~\citep{wong2021leveraging}, to improve the robustness out of distribution and uncertainty performance~\citep{zhang2021subnetwork}, to detect non-spurious correlation~\citep{sagawa2020investigation}, etc.

% By performing an over-parameterization in the space-time manifold, a randomly initialized sparse network can outperform the highly over-parameterized dense model by a large margin. 
% Based on In-Time Over-Parameterization, we better understand the underlying mechanism of Dynamic Sparse Training and

% and find that as long as there are sufficient parameters that have been reliably explored, Dynamic Sparse Training can at least match, and in most cases exceed the dense performance. The results

% Inspired by In-Time Over-Parameterization, we improve state-of-the-art sparse training performance with ResNet-50 on Imagenet by encouraging a reliable and reliable sparse connectivity exploration. We further compare different sparsity-inducing methods both with a typical training time and an extended training time. The sparse models with sufficient and reliable parameter exploration consistently achieve better performance over those with specially-designed sparse initialization. 

% \textcolor{blue}{The limitation of this work is we have to choose proper $\Delta T$}

\bibliography{example_paper}
\bibliographystyle{icml2021}

\clearpage

\onecolumn
\appendix
\appendixpage
% \appendixpage
\section{Experimental Settings of Hypothesis Evaluation}

\label{Imp_settings}
In this Appendix, we describe the experimental settings of the hypothesis evaluation in Section \ref{sec:hy_eva}.
\subsection{Models}
We use MLP on CIFAR-10, VGG-16 on CIFAR-10, ResNet-34 on CIFAR-100 to work through our hypothesis. We describe these models in detail as follows: 

\textbf{MLP.} MLP is a clean three-layer MLP with ReLU activation for CIFAR-10. The number of neurons of each layer is 1024, 512, 10, respectively. No other regularization such as dropout or batch normalization is used further. 

\textbf{VGG-16.} VGG-16 is a modified CIFAR-10 version of the original VGG model introduced by \citet{lee2018snip}. The size of the fully-connected layer is reduced to 512 and the dropout layers are replaced with batch normalization to avoid any other sparsification.

\textbf{ResNet-34.} ResNet-34 is the CIFAR-100 version of ResNet with 34 layers introduced by \citet{he2016deep}.

\subsection{Algorithm}
We choose Sparse Evolutionary Training (SET) \citep{mocanu2018scalable} as the DST method to evaluate our hypothesis. SET helps to avoid the dense over-parameterization bias introduced by the gradient-based methods e.g., RigL and SNFS, as the latter utilize dense gradients in the backward pass to explore new weights. SET starts from a random sparse topology (\textit{Erd{\H{o}}s-R{\'e}nyi}), and optimize the sparse connectivity towards a scale-free topology during training. 

This algorithm contains three key steps:
\begin{enumerate}
   \item Initializing a sparse neural network with \textit{Erd{\H{o}}s-R{\'e}nyi} random graph at a sparsity of S.
   \item Training the sparse neural network for $\Delta T$ iterations.
   \item Removing weights according to the standard magnitude pruning and growing new weights in a random fashion. 
\end{enumerate}
Steps 2 and 3 will be repeated iteratively until the end of the training. By doing this, SET maintains a fixed parameter count throughout training. 

\subsection{Training}
\begin{table*}[ht!]
\small
\centering
\caption{Experiment hyperparameters of the hypothesis evaluation in Section \ref{sec:hy_eva}. The hyperparameters include Learning Rate (LR), Batch Size (BS), Typical Training Epochs (TT Epochs), Learning Rate Drop (LR Drop), Weight Decay (WD), Sparse Initialization (Sparse Init), Update Interval of the Extended Training ($\Delta T$), Pruning Rate Schedule (Sched), Initial Pruning Rate (P), etc.}
\label{tab:hypo_hyper}
\begin{tabular}{cccccccccccc}
\toprule
Model & Data & Methods & LR & BS & TT Epochs & LR Drop &  WD & Sparse Init &  $\Delta T$ & Sched & P   \\
\midrule
MLP & CIFAR-10 & RigL & 0.01 & 128  & 200 & 10x & 5e-4 & ER & 4000 & Cosine & 0.5\\
MLP & CIFAR-10 & SET & 0.01 & 128  & 200 & 10x & 5e-4 & ER & 1500 & Cosine & 0.5\\
VGG-16 & CIFAR-10 & SET & 0.1 & 128  & 250 & 10x & 5e-4 & ERK & 2000 & Cosine & 0.5\\
ResNet-34 & CIFAR-100 & SET & 0.1 &  128 & 200 & 10x & 1e-4 & ERK & 1000 & Cosine & 0.5\\
\bottomrule
\end{tabular}
\end{table*}

We basically follow the experimental settings from \citet{dettmers2019sparse}. 

For models trained for a typical time, we train them with various update interval $\Delta T$ reported in Figure \ref{fig:typical_time}. We use a set of 10\% training data as the validation set and train on the remaining training data. Weight growth is guided by random sampling and weight pruning is guided by magnitude. We do not specifically finetune the starting point and the finishing point of the parameter exploration. The exploring operation is performed throughout training. The initial sparse connectivity is sampled by the \textit{Erd{\H{o}}s-R{\'e}nyi} distribution introduced in \citet{mocanu2018scalable}. We set the initial pruning rate as 0.5 and gradually decay it to 0 with a cosine annealing, as introduced in \citet{dettmers2019sparse}. The remaining training hyperparameters are set as follows:

\textbf{MLP.} We train sparse MLPs for 200 epochs by momentum SGD with a learning rate of 0.01 and a momentum coefficient of 0.9. We use a small learning rate 0.01 rather than 0.1, as the dense MLP doesn't converge with a learning rate of 0.1. We decay the learning rate by a factor of 10 every 24000 iterations. We set the batch size as 128. The weight decay is set as 5.0e-4. 

\textbf{VGG-16.} We strictly follow the experimental settings from \citet{dettmers2019sparse} for VGG-16. All sparse models are trained with momentum SGD for 250 epochs with a learning rate of 0.1, decayed by 10 every 30000 mini-batches. We use a batch size of 128 and weight decay to 5.0e-4. 

\textbf{ResNet-34.} We train sparse ResNet-34 for 200 epochs with momentum SGD with a learning rate of 0.1, decayed by 10 at the 100 and 150 epoch. We use a batch size of 128 and weight decay to 1.0e-4. 

For models trained for an extended training time, we simply extend the training time and the anchor epochs of the learning rate schedule, while using a large $\Delta T$. The update interval $\Delta T$ is chosen according to the trade-off shown in Figure \ref{fig:typical_time}. Besides the learning steps, the anchor epochs of the learning rate schedule and the pruning rate schedule are also scaled by the same factor. For each training time, the accuracy are averaged over 3 seeds with mean and standard deviation. More detailed training hyperparameters are shared in Table \ref{tab:hypo_hyper}.

\section{Implementation Details of RigL-ITOP in Section \ref{sec:impro_dst}}
\label{App_riglITOP}

\begin{table*}[ht!]
\small
\centering
\caption{Experiment hyperparameters in Section \ref{sec:impro_dst} and Section \ref{sec:comparison}. The hyperparameters include Learning Rate (LR), Batch Size (typical training time / extended training time) (BS), Training Epochs (typical training time / extended training time) (Epochs), Learning Rate Drop (LR Drop), Weight Decay (WD), Sparse Initialization (Sparse Init), Update Interval ($\Delta T$), Pruning Rate Schedule (Sched), Initial Pruning Rate (P), etc.}
\label{tab:ITOP}
\begin{tabular}{cccccccccccc}
\toprule
Model & Data & Methods & LR & BS & Epochs & LR Drop &  WD & Sparse Init & $\Delta T$ & Sched & P   \\
\midrule
MLP & CIFAR-10 & SET-ITOP & 0.01 & 32 / 128 & 200 / 4000 & 10x & 5e-4 & ER & 1500 & Cosine & 0.5\\
MLP & CIFAR-10 & RigL-ITOP & 0.01 & 32 / 128 & 200 / 4000 & 10x & 5e-4 & ER & 4000 & Cosine & 0.5\\
ResNet-34 & CIFAR-100 & SET-ITOP & 0.1 & 32 / 128 & 200 / 4000 & 10x  & 1e-4 & ERK & 1500 & Cosine & 0.5\\
ResNet-34 & CIFAR-100 & RigL-ITOP & 0.1 & 32 / 128 & 200 / 4000 & 10x  & 1e-4 & ERK & 4000 & Cosine & 0.5\\
ResNet-50 & ImageNet & RigL-ITOP & 0.1 & 64 / - & 100 / - & 10x & 1e-4 & ERK & 4000 & Cosine & 0.5\\

\bottomrule
\end{tabular}
\end{table*}

In this Appendix, we describe our replication of RigL \citep{evci2019rigging} and the hyperparameters we used for RigL-ITOP. 

RigL is a state-of-the-art DST method growing new weights that are expected to receive gradient with high magnitude in the next iteration. Besides, it shows the proposed sparse distribution \textit{Erd{\H{o}}s-R{\'e}nyi-Kernel} (ERK) improves the sparse performance over the \textit{Erd{\H{o}}s-R{\'e}nyi} (ER). Since RigL is originally implemented with TensorFlow, we replicate it with PyTorch based on the implementation from \citet{dettmers2019sparse}. We note that RigL tunes the starting epoch and the ending point of the mask update. To encourage more exploration, we do not follow this strategy and explore sparse connectivities throughout training. We train sparse ResNet-50 for 100 epochs, the same as \citet{dettmers2019sparse,evci2019rigging}. The learning rate is linearly increased to 0.1 with a warm-up in the first 5 epochs and decreased by a factor of 10 at epochs 30, 60, and 90. To reach a high and reliable In-Time Over-Parameterization rate, we use a small batch size of 64 and an update interval of 4000. Batch sizes lower than 64 lead to worse test accuracy. ImageNet experiments were run on 2x NVIDIA Tesla V100. With more fine-tuning, the results of RigL-ITOP (e.g. extended training time) can likely be improved, but we lack the resources to do it. We share the hyperparameters of RigL-ITOP in Table \ref{tab:ITOP}.

\section{Implementation Details in Section \ref{sec:comparison}}
\label{Rep_other_methods}
In this Appendix, we describe the hyperparameters of SET-ITOP used in~\ref{sec:comparison} in Table \ref{tab:ITOP}. The replication details of LTH and SNIP are given below.

\textbf{LTH.} Lottery Ticket Hypothesis (LTH) \citep{frankle2018the} shows that there exist sub-networks that can match the accuracy of the dense network when trained with their original initializations. We follow the PyTorch implementation provide by \citet{liu2019rethinking} on GitHub\footnote[1]{\url{https://github.com/Eric-mingjie/rethinking-network-pruning}} to replicate LTH.

Give the fact that the iterative pruning process of LTH would lead to much larger training resource costs than SNIP and static sparse training, we use one-shot pruning for LTH. For the typical training time setting, we first train a dense model for 200 epochs, after which we use global and one-shot magnitude pruning to prune the model to the target sparsity and retrain the pruned model with its original initializations for 200 epochs. 

\textbf{SNIP.} Single-shot network pruning (SNIP) proposed in \citet{lee2018snip}, is a method that attempts to prune at initialization before the main training based on the connection sensitivity score $s_i=|\frac{\partial L}{\partial w_i}w_i|$. The weights with the smallest score are pruned. We replicate SNIP based on the PyTorch implementation on GitHub\footnote[1]{\url{https://github.com/mil-ad/snip}}. Same as \citet{lee2018snip}, we use a mini-batch of data to calculate the important scores and obtain the sparse model in a one-shot fashion before the initialization. After that, we train the sparse model without any sparse exploration for 200 epochs.

% \textbf{GMP.} Gradual magnitude pruning (GMP) was first proposed in \citet{zhu2017prune}, later studied by \citet{gale2019state}. Although starting from a dense model, the model can be gradually pruned early in training. We use the PyTorch implementation provided by \citet{kusupati2020soft} on GitHub\footnote[1]{\url{https://github.com/RAIVNLab/STR}}. For the typical training time setting, we start the pruning process at the 40 epoch and the sparse model with target sparsity is obtained at the end of training. For the extended training time setting, we first train the dense model for 800 epochs after which the pruning starts until the end of the training. 

\section{Extended Training Performance of RigL with $\Delta T = 1500$}
\label{app:rigl_1500}
\begin{figure*}[ht]
    \centering
    \includegraphics[width=\textwidth]{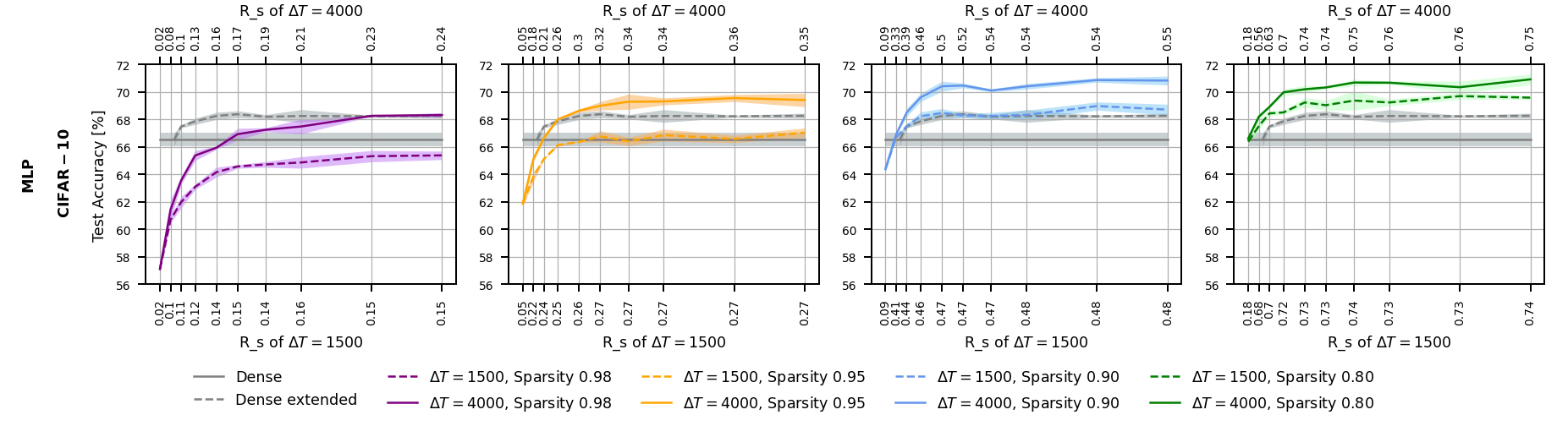}
    \caption
    {\small Extended training performance of RigL with update interval $\Delta T = 1500$ and $\Delta T = 4000$.}  
    \label{Fig:comp_rigl}
\end{figure*}

According to the results from Figure \ref{Fig:rigl}, we can see the $\Delta T=4000$ is a good choice for the update interval of RigL. What if we choose a small update interval, e.g., $\Delta T=1500$? Here we compare the extended training performance of RigL with two different update intervals 1500 and 4000. The results are shown in Figure \ref{Fig:comp_rigl}. It is clear to see models trained with $\Delta T=1500$ fall short of models trained with $\Delta T=4000$, which indicates small update intervals is not sufficient for newly weights to catch up the existing weights in terms of magnitude. More importantly, although expected to perform sparse exploration more frequently, models trained with $\Delta T=1500$ end up with a lower $R_s$ than the ones trained with $\Delta T=4000$. These results highlight the importance of the sufficient training time for the new weights. 

\section{Test Accuracy of RigL with Various Batch Sizes}
\label{app:rigl_bs}
\begin{figure*}[h]
    \centering
    \includegraphics[width=\textwidth]{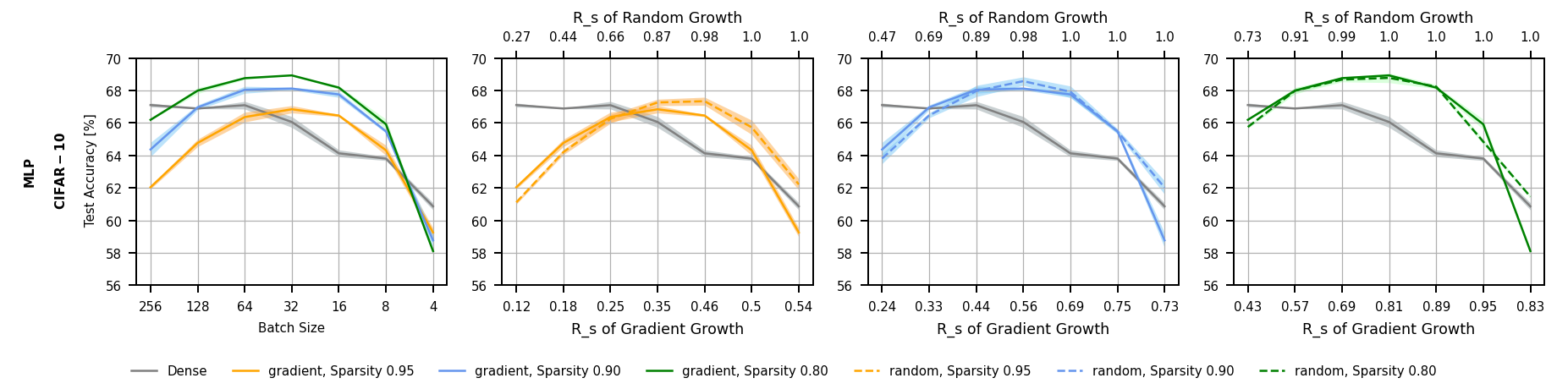}
    \caption
    {\small Test accuracy of RigL with various batch sizes. The update interval $\Delta T$ is set as 4000.}  
    \label{Fig:various_bs_rigl}
\end{figure*}

In this Appendix, we evaluate the performance of RigL with different batch sizes. We choose MLP as our model and the update interval $\Delta T = 4000$. The results are shown in Figure \ref{Fig:various_bs_rigl}. Similar with SET, the performance of RigL also increases as the batch size decrease from 256 to 32. After that, the performance starts to drop due to the noisy input caused by the extreme small batch sizes. The In-Time Over-Parameterization rate ($R_s$) of RigL is again bounded up to some values. We also provide the comparison between RigL (solid lines) and SET (dashed lines) in this setting. We find a similar pattern with the extended training time, that is, RigL outperforms SET when $R_s$ is small but falls short of SET when sufficient parameters have been reliably explored.

\section{Regrowing from the Non-Activated Weights First}
\label{app:samping}

One direct way to increase the In-Time Over-Parameterization rate during a typical training time is to sample from the non-activated weights first when performing weight growing. We evaluate this idea with SET by regrowing the non-activated weights first and report the results as SET+ with $(\textrm{mean}\pm\textrm{std}, R_s)$ in Table~\ref{tab:sampling}. We training sparse ResNet-18 on CIFAR-10 for 250 epochs with a learning rate of 0.1 decayed by 10x at 125, 187 epochs, a batch size of 128, a pruning rate of 0.5.

When the parameter exploration is insufficient (small $R_s$), SET+ consistently achieves higher accuracy and higher $R_s$ than SET. Whiling effective, the $R_s$ increase achieved by this modification is relatively limited. This observation highlights an important
direction for future work to achieve high $R_s$ within a typical training time.

\begin{table*}[h]
\scriptsize
\vspace{-0.3em}
\centering
\caption{Performance of sparse ResNet-18 on CIFAR-10 with various pruning rates. The results are run three times and reported with $(\textrm{mean}\pm\textrm{std}, R_s)$. The highest test accuracies are marked in bold.}
\label{tab:sampling}
\resizebox{1.0\textwidth}{!}{
\begin{tabular}{c c  c| c c ccc}
\toprule
Sparsity &  Method & P &$\Delta T=15000$ & $\Delta T=10000 $  & $\Delta T=8000 $ & $\Delta T=5000 $ & $\Delta T=3000 $  \\ 
\midrule
\multirow{2}{*}{0.9} & SET & 0.5 & (94.30 $\pm$ 0.16, 0.162) & (94.47 $\pm$ 0.14, 0.201) & (94.25 $\pm$ 0.10, 0.228) & (94.36 $\pm$ 0.08, 0.302) & ({\bf 94.54} $\pm$ 0.05, 0.411) \\
& SET+ & 0.5 &  ({\bf 94.43} $\pm$ 0.14, 0.169) & ({\bf 94.59} $\pm$ 0.11, 0.215) & ({\bf 94.54} $\pm$ 0.28, 0.247) & ({\bf 94.38} $\pm$ 0.07, 0.342) & (94.53 $\pm$ 0.03, 0.492)  \\ 
\midrule
\multirow{2}{*}{0.95} & SET & 0.5 & (93.57 $\pm$ 0.16, 0.086) & (93.46 $\pm$ 0.04, 0.108) & ({\bf 93.67} $\pm$ 0.04, 0.124) & (93.60 $\pm$ 0.04, 0.170) & (93.61 $\pm$ 0.09, 0.241) \\ 
& SET+ & 0.5 & ({\bf 93.66} $\pm$ 0.14, 0.088) & ({\bf 93.70} $\pm$ 0.03, 0.114) & (93.66 $\pm$ 0.15, 0.133) & ({\bf 93.78} $\pm$ 0.04, 0.186) & ({\bf 94.00} $\pm$ 0.08, 0.272) \\
\bottomrule
\end{tabular}}
\vspace{-0.3cm}
\end{table*}
% \appendix
% \section{Do \emph{not} have an appendix here}

% \textbf{\emph{Do not put content after the references.}}
% %
% Put anything that you might normally include after the references in a separate
% supplementary file.

% We recommend that you build supplementary material in a separate document.
% If you must create one PDF and cut it up, please be careful to use a tool that
% doesn't alter the margins, and that doesn't aggressively rewrite the PDF file.
% pdftk usually works fine. 

% \textbf{Please do not use Apple's preview to cut off supplementary material.} In
% previous years it has altered margins, and created headaches at the camera-ready
% stage. 
%%%%%%%%%%%%%%%%%%%%%%%%%%%%%%%%%%%%%%%%%%%%%%%%%%%%%%%%%%%%%%%%%%%%%%%%%%%%%%%
%%%%%%%%%%%%%%%%%%%%%%%%%%%%%%%%%%%%%%%%%%%%%%%%%%%%%%%%%%%%%%%%%%%%%%%%%%%%%%%

\end{document}